\documentclass[runningheads]{llncs}
\usepackage{graphicx}
\usepackage{cite}
\usepackage{amsmath,amssymb,amsfonts}
\usepackage{algorithmic}
\usepackage{graphicx}
\usepackage{textcomp}
\usepackage{xcolor}
\usepackage{booktabs}   % \toprule and \bottomrule in tables
\usepackage{array}	% for defining fixed-width columns in tables
\usepackage{mathabx} % for down-right arrow (\drsh)
\usepackage{todonotes}     % options [disable]

\usepackage{hyperref} % clickable links and reference
\hypersetup{
    colorlinks=true,
    linkcolor=blue,
    filecolor=cyan,      
    urlcolor=blue,
    citecolor=cyan,
}

\usepackage{tikz}
\newcommand{\myarrow}[1][]{%
  \begin{tikzpicture}[#1]%
    \draw (0,0.7ex) -- (0,0) -- (0.75em,0);
    \draw (0.55em,0.2em) -- (0.75em,0) -- (0.55em,-0.2em);
  \end{tikzpicture}%
}

\graphicspath{{../../graphics/}}

\begin{document}
\title{Avoiding Implementation Pitfalls of ``Matrix Capsules with EM Routing'' by Hinton \textit{et al.}}
\titlerunning{Avoiding Implementation Pitfalls of Matrix Capsules}
% If the paper title is too long for the running head, you can set
% an abbreviated paper title here
%
\author{Ashley Daniel Gritzman\inst{1}\orcidID{0000-0002-9949-157X}}
\authorrunning{A. Gritzman}
% First names are abbreviated in the running head.
% If there are more than two authors, 'et al.' is used.
%
\institute{IBM Research, Johannesburg, South Africa\\
\email{ashley.gritzman@za.ibm.com}}
\maketitle              % typeset the header of the contribution
\begin{abstract}
	
	% Adapted for HBAI
% If the paper title is too long for the running head, you can set
	The recent progress on capsule networks by Hinton \textit{et al.} has generated considerable excitement in the machine learning community. The idea behind a capsule is inspired by a cortical minicolumn in the brain, whereby a vertically organised group of around 100 neurons receive common inputs, have common outputs, are interconnected, and may well constitute a fundamental computation unit of the cerebral cortex. However, Hinton's paper on ``Matrix Capsule with EM Routing'' was unfortunately not accompanied by a release of source code, which left interested researchers attempting to implement the architecture and reproduce the benchmarks on their own. This has certainly slowed the progress of research building on this work. While writing our own implementation, we noticed several common mistakes in other open source implementations that we came across. In this paper we share some of these learnings, specifically focusing on three implementation pitfalls and how to avoid them: (1) parent capsules with only one child; (2) normalising the amount of data assigned to parent capsules; (3) parent capsules at different positions compete for child capsules. While our implementation is a considerable improvement over currently available implementations, it still falls slightly short of the performance reported by Hinton \textit{et al.} (2018). The source code for this implementation is available on GitHub at the following URL: \url{https://github.com/IBM/matrix-capsules-with-em-routing}.

\end{abstract}
\section{Introduction}

	Geoffrey Hinton has been talking about ``capsule networks'' for a long time, so when his team published their recent progress on this topic, it naturally created quite a stir in the machine learning community. The idea behind a capsule is inspired by a cortical minicolumn in the brain, whereby a vertically organised group of around 100 neurons receive common inputs, have common outputs, are interconnected, and may well constitute a fundamental computation unit of the cerebral cortex \cite{Cruz2005}. In the context of machine learning, a capsule is a group of neurons whose outputs represents not only the probability that an entity exists, but also different properties of the same entity. Capsules may encode information such as orientation, scale, velocity, and colour. Layers in a capsule network learn to assemble these entities to form parts of a larger whole. 

	% Added to try make this work more relevant to HBAI workshop [http://www.ijcai-hbai.org/]
	% Source: https://medium.com/ai%C2%B3-theory-practice-business/understanding-hintons-capsule-networks-part-i-intuition-b4b559d1159b
	In computer graphics, a scene is represented in abstract form comprising objects and their corresponding instantiation parameters (e.g. x, y location, and angle). A rendering function then converts this abstract representation into an image. Hinton argues that the brain does `inverse graphics'~\cite{Hinton2012}, which essentially means deconstructing visual information received by the eyes into a hierarchical representation of the world, and then trying to match it with already learned patterns and relationships stored by the brain. A capsule network is basically a neural network that tries to perform inverse graphics.
	
	Hinton \textit{et al.}~\cite{Hinton2011} first introduced the concept of capsule networks in 2011 when they used a transformation matrix in a ``transforming auto-encoder'' that learned to transform a stereo pair of images into a stereo pair from a slightly different viewpoint. But it was only towards the end of 2017 that Sabour \textit{et al.}~\cite{Sabour2017} published a capsule network architecture featuring dynamic routing-by-agreement that managed to reach state-of-the-art performance on MNIST~\cite{LeCun1998}, and considerably better results than CNNs on MultiMNIST~\cite{Sabour2017} (a variant with overlapping pairs of different digits). Then in 2018, Hinton \textit{et al.}~\cite{Hinton2018} published ``Matrix Capsules with EM Routing'' to address some of the deficiencies of Sabour \textit{et al.}~\cite{Sabour2017}, and reported a reduction in the test error on smallNORB~\cite{LeCun2004} of $45\%$ compared to state-of-the-art.

	The matrix capsule version of a capsule network is described as follows \cite{Hinton2018}: ``each capsule has a logistic unit to represent the presence of an entity and a 4x4 matrix which could learn to represent the relationship between that entity and the viewer (the pose). A capsule in one layer votes for the pose matrix of many different capsules in the layer above by multiplying its own pose matrix by trainable viewpoint-invariant transformation matrices that could learn to represent part-whole relationships. Each of these votes is weighted by an assignment coefficient. These coefficients are iteratively updated for each image using the Expectation-Maximization algorithm such that the output of each capsule is routed to a capsule in the layer above that receives a cluster of similar votes.''

	Sabour \textit{et al.}~\cite{Sabour2017} made their code for ``Dynamic Routing between Capsules'' available on GitHub~\cite{Sabour2018}, but unfortunately Hinton \textit{et al.}~\cite{Hinton2018} did not do the same for ``Matrix Capsules with EM Routing'', which may somewhat explain the slower progress in research building on this work. While implementing this work ourselves, we noticed several common mistakes in other implementations that we came across, and also discovered a couple of pitfalls which may prevent the network from operating as intended.

	In this paper we share some of these learnings, specifically focusing on three implementation pitfalls and how to avoid them: (1) parent capsules with only one child; (2) normalising the amount of data assigned to parent capsules; (3) parent capsules at different positions compete for child capsules.

	To make this paper slightly easier to consume for readers not entirely familiar the work, we try to simplify the terminology where possible, for example we use the terms ``child capsules'' to represent capsules in lower layer $L$, and ``parent capsules'' to represent capsules in higher layer $L+1$. However, we do assume that the reader is already familiar with Hinton's paper \cite{Hinton2018}, which is necessary in order to understand the discussion on the pitfalls that follows below.

\section{Understanding and Avoiding Pitfalls}

	\subsection{Parent Capsules with Only One Child}
		
\begin{figure}[htbp]
			\centerline{\includegraphics[width=1\columnwidth]{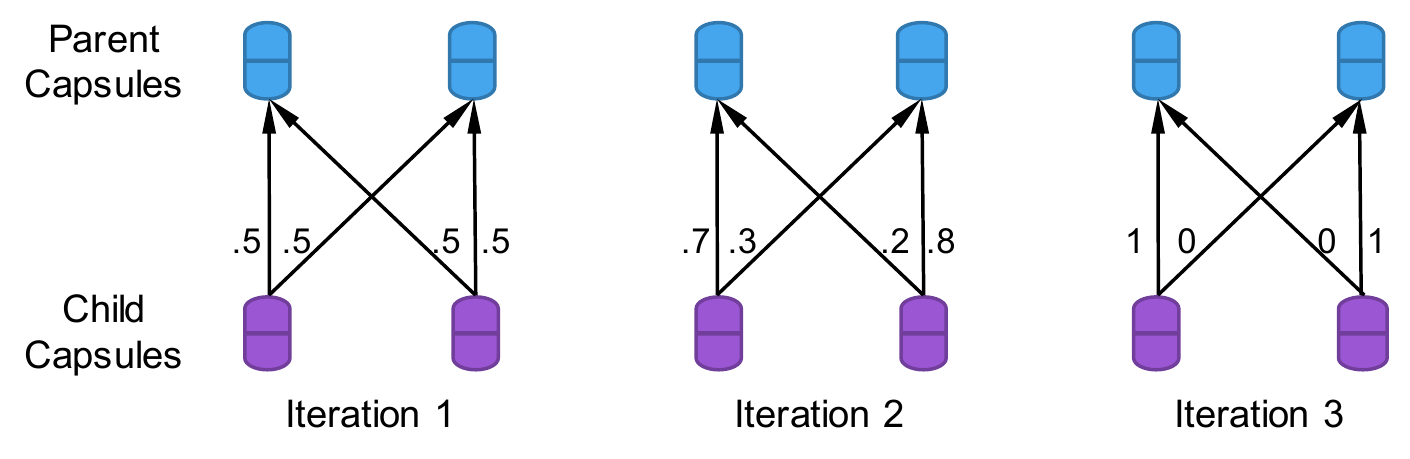}}
			\caption{Illustration of assignment probabilities between two capsule layers over three iterations of EM routing. At iteration 3 each parent capsule receives input from only one child capsule.}
			\label{fig:one_to_one}
		\end{figure}

		The EM routing algorithm is a form of cluster finding which iteratively adjusts the assignment probabilities between child capsules and parent capsules. Fig.~\ref{fig:one_to_one} illustrates this process: at the start of EM routing, the output of each child capsule is evenly distributed to all of the parent capsules. As the EM algorithm proceeds, the affinity of parent capsules for particular child capsules increases, and eventually each child capsule may contribute to only one parent capsule. This phenomenon does not cause a problem if each parent capsule comprises multiple child capsules, however the situation may arise whereby a parent capsule comprises only one child capsule. This situation is similar to a clustering scenario whereby a cluster has only one data point. Since the EM routing algorithm fits a Gaussian distribution to each of the parent capsules, it is necessary to calculate the mean $\mu_j^h$ and variance $(\sigma_j^h)^2$ of each parent capsule. If the parent capsule comprises only one child capsule, then the variance of the parent capsule is 0. This causes numerical instability when calculating the activation cost in Eq~(\ref{eq:cost}), as $log(\sigma_j^h)$ is undefined at $\sigma_j^h = 0$. Furthermore, a Gaussian distribution with a variance of 0 is the unit impulse centered at the mean $\mu_j^h$, so $p_j(\mu_j^h) = \infty$ which causes numerical overflow. 

		\begin{equation}
			cost^h \leftarrow \big(\beta_u + log(\sigma_j^h)\big)\sum_{i}{R_{ij}}
			\label{eq:cost}	
		\end{equation}

		This problem of parent capsules having only one child capsule occurs more frequently as the number of routing iterations increases, whereby the assignment probabilities tend to either 0 or 1. In our experiments with the \textit{smaller} capsule network configuration of $A=64$, $B=8$, $C=D=16$, the problem did not occur during training with one or two iterations, but occurred consistently for $iteratations \geq 3$.

		Furthermore, the occurrence of this problem also depends on the ratio of child capsules to parent capsules. If the ratio is high, meaning many child capsules feeding fewer parent capsules, then the problem only occurs at higher routing iterations. Whereas, if the ratio is low and approaches $1{:}1$, or even lower (i.e. more parent capsules than child capsules), then the problem starts occurring at a lower number of routing iterations.  

		To address this problem in our implementation, we impose a lower bound on the variance $(\sigma_j^h)^2$ by adding $\epsilon = 10^{-4}$.

	\subsection{Normalising the Amount of Data Assigned to Parent Capsules}

		In Eq~(\ref{eq:cost}) computing the activation cost, $\sum_{i}{R_{ij}}$ is the amount of data assigned to parent capsule $j$ from all child capsules. 

		For the convolutional capsule layers, each child capsule within the kernel feeds to one spatial location in the parent layer containing \textit{O} types of capsules. If all child capsules within a convolutional kernel assign their data to only one parent capsule type, then the maximum data that a parent capsule can receive is give by: 
		\begin{equation}
			max\_data = K \times K \times I
			\label{eq:max_data}
		\end{equation}
		where $K$ is the kernel size, and $I$ is the number of input capsule types.

		The mean data received by parent capsules (assuming that all child capsules are active) is the total number of child capsules divided by the total number of parent capsules: 
		\begin{equation}
			mean\_data = \frac{child\_W \times child\_H \times I}{parent\_W \times parent\_H \times O}
			\label{eq:mean_data}
		\end{equation}
		where $W$ and $H$ denote the spatial width and height of the tensors containing the capsules, and $O$ is the number of output capsule types.

		For the final output layer denoted \textit{class\_caps}, the spatial dimensions of the child tensor is flattened such that the child capsules are fully connected to the \textit{class} capsules. 

		Table~\ref{tab:layers} shows a summary of the layers in the \textit{smaller} capsule network configuration. For capsule layers connected with EM routing, the maximum and mean assignment data is calculated with equations~(\ref{eq:max_data}) and (\ref{eq:mean_data}). The mean assignment data is similar for both the \textit{conv\_caps1} and \textit{conv\_caps2} layers at $2.61$ and $1.96$ respectively, however notice that the mean data of the \textit{class\_caps} layer is ${\approx}30 \times$ larger at $80.0$. The larger assignment data for the \textit{class\_caps} layer occurs since each parent capsule in this layer is fully connected to all child capsules in \textit{conv\_caps2} layer. Therefore, the $5 \times 5 \times 16=400$ capsules in the \textit{conv\_caps2} layer feed to just $5$ capsules in the \textit{class\_caps} layer.

 		% TABLE: EM routing data
 		\begin{table}[hbt]
 			\centering
 			\caption{Summary of layers in the the \textit{smaller} capsule network configuration ($A=64$, $B=8$, $C=D=16$) showing the maximum and mean assignment data between layers with EM routing. \textit{K} is the kernel size, \textit{S} is the stride, \textit{Ch} is the number of channels in a regular convolution, \textit{I} is the number of input capsule types, \textit{O} is the number of output capsule types, \textit{W} and \textit{H} are the spatial width and height. The \textit{Output shape} shows the dimensions of the tensor containing the \textit{activations (batch size, W, H, Ch~or~O)}; the tensor containing the \textit{poses} would have additional dimensions 4x4 for the \textit{pose} matrix.}
 			\label{tab:layers}

 			% \newcolumntype{C}[1]{>{\centering}m{#1}}
 			\newcolumntype{L}[1]{>{\raggedright\let\newline\\\arraybackslash\hspace{0pt}}p{#1}}
			\newcolumntype{C}[1]{>{\centering\let\newline\\\arraybackslash\hspace{0pt}}p{#1}}
			\newcolumntype{R}[1]{>{\raggedleft\let\newline\\\arraybackslash\hspace{0pt}}p{#1}}

 			\begin{tabular}[t]{l l l c c}
 				\toprule
 					& & & \multicolumn{2}{l}{EM Rt. Data} \\
						\cline{4-5}	
			 		Layer	& Details & Output shape & Max & Mean \\
 				\midrule
 					% myarrow~
 					input 	&  & (?, 32, 32, 1) \\
 					{\myarrow}relu\_conv1	&  K=5, S=2, Ch=64 & (?, 16, 16, 64) \\
 					{\myarrow}primary\_caps & K=1, S=1, Ch=8 & (?, 16, 16, 8) \\
					{\myarrow}conv\_caps1 & K=3, S=2, O=16 & (?, 7, 7, 16) & 72 & 2.61 \\
					{\myarrow}conv\_caps2 & K=3, S=1, O=16 & (?, 5, 5, 16) & 144 & 1.96 \\ 
					{\myarrow}class\_caps & flatten, O=5 & (?, 1, 1, 5) & 400 & 80.0 \\
 				\bottomrule
 			\end{tabular}
 		\end{table}	

 		We now consider the effects of the large discrepancy in assignment data between the \textit{class\_caps} layer and the other layers. While the mean data was calculated under the assumption that all child capsules are active (which is unlikely), nevertheless the actual data assigned to a parent capsule $\sum_{i}{R_{ij}}$ will be proportional to the mean value of the layer.

 		Consider the computation of the parent activations from the M-step of Procedure~1 in \cite{Hinton2018}:
		\begin{equation*}
			a_j \leftarrow logistic\Bigg(\lambda\bigg(\beta_a - \sum_{h}{\Big(\beta_u + log(\sigma_j^h)\Big)}\sum_{i}{R_{ij}}  \bigg) \Bigg)
			% \label{eq:activ_1}	
		\end{equation*}

		\noindent Since $\beta_u$ is per capsule type and does not depend on $h$:
		\begin{equation}
			a_j \leftarrow logistic\Bigg(\lambda\bigg(\beta_a - \sum_{i}{R_{ij}} \Big(H\beta_u + \sum_{h}{log(\sigma_j^h)\Big)} \bigg) \Bigg)
			\label{eq:activ_2}	
		\end{equation}

		\noindent Finally:
		\begin{equation}
			a_j \leftarrow logistic\Bigg(\lambda\beta_a - \lambda\sum_{i}{R_{ij}} \Big(H\beta_u + \sum_{h}{log(\sigma_j^h)\Big)} \Bigg)
			\label{eq:activ_3}	
		\end{equation}

		Consider the second term in Eq~\ref{eq:activ_2}, notice that the cost of activating a parent capsule is scaled by the total amount of data received by that capsule $\sum_{i}{R_{ij}}$. In Eq~(\ref{eq:activ_3}) the first term $\lambda\beta_a$ sets the operating point on the logistic curve, and the second term determines the perturbations about this point. If the $\sum_{i}{R_{ij}}$ scaling is too small, then all the output activations will not deviate from the operating point. But if the $\sum_{i}{R_{ij}}$ scaling is too large, then all the parent capsules will either be fully active or inactive. The most desirable situation occurs when the input to the logistic function is nicely distributed over a useful range (e.g. $[-5, 5]$), so the output of the logistic function is nicely distributed over the range $[0, 1]$.

		The impact of the ${\approx}30 \times$ difference in assignment data is that the range of the input to the logistic function will not be distributed over a useful range for all capsule layers. The values $\beta_a$ and $\beta_v$ are learned discriminatively for each layer, and can to some extent compensate for this effect. However, since $\beta_a$ and $\beta_v$ are initialised randomly, the problem will be more pronounced at the start of training and may prevent the network from learning. 

		This problem can be addressed to an extent by carefully initialising $\beta_a$ and $\beta_v$, which will need to be different for the \textit{class\_caps} layer and the preceding \textit{conv\_caps1} and \textit{conv\_caps2} layers. It will also be necessary to ensure that the $\lambda$ scaling value ensures a useful range for the input to the logistic function at every layer. In our implementation we adopt a different approach, and instead scale the amount of data assigned to a parent capsule $(\sum_{i}{R_{ij}})$ relative to the mean data in a particular layer (see Table~\ref{tab:layers} for scaling values). We find this approach to be more robust to the initial values of $\beta_a$ and $\beta_v$. 

	\subsection{Parent Capsules at Different Positions Compete for Child Capsules}

\begin{figure}[htbp]
			\centerline{\includegraphics[width=0.95\columnwidth]{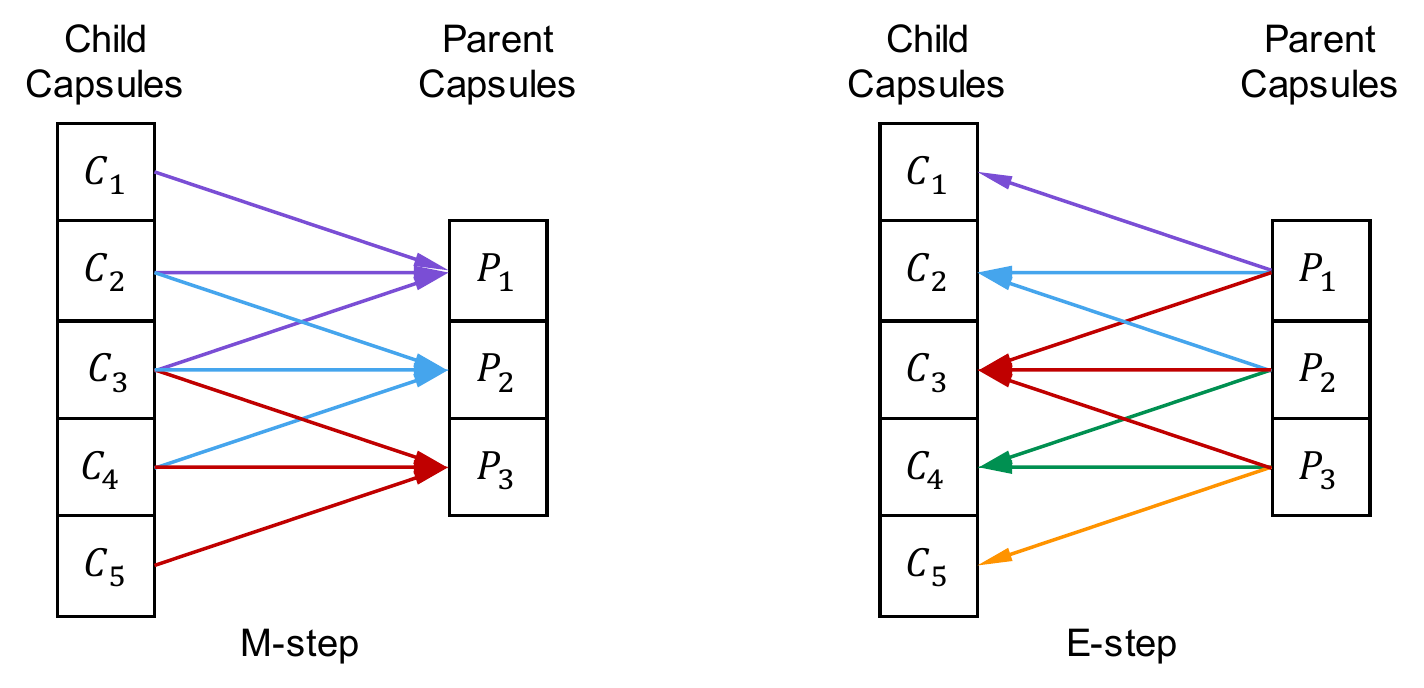}}
			\caption{Connectivity between parent capsules and child capsules resulting from 1D convolution with a kernel size of 3 and a stride of 1. In the \textbf{M-step}, all parent capsules receive input from 3 child capsules; in \textbf{E-step}, child capsules towards the edges receive feedback from fewer parent capsules, while capsules at towards the center receive feedback from up to 3 parent capsules.}
			\label{fig:em_connections}
		\end{figure}

		Consider the case of a 1D convolutional capsule layer, with a kernel size of $3$ and a stride of $1$, and where both child and parent layers contain only $1$ capsule type. In the M-step shown in Fig.~\ref{fig:em_connections}, the kernel slides over the child capsules resulting in each parent receiving votes from 3 child capsules. In the E-step, child capsules towards the edges receive feedback from fewer parent capsules, while capsules towards the center receive feedback from up to $K$ parent capsules. In the example of Fig.~\ref{fig:em_connections}, $\{C_1, C_5\}$ receive feedback from one parent capsule each, $\{C_2, C_4\}$ each receive feedback from two parent capsules, and $\{C_3\}$ receives feedback from three parent capsules. 

		It is clear from Fig.~\ref{fig:em_connections} that child capsules receive feedback from parent capsules at different spatial positions, and therefore these parent capsules must compete for the vote of the child capsule. The competition happens in the update of the assignment probabilities in the E-step, where we normalise across all parent capsules competing for a particular child capsule. This point was further clarified by the paper authors in response to a question on OpenReview.net \cite{Calvano2018}.

		We reviewed several open source implementations on GitHub, and found that incorrect normalisation in the E-step is a common mistake. In particular, the implementations normalise only across parent capsule types, and not across parent capsule positions. This has the unintended effect of preventing parent capsules at different positions from competing for child capsules. The correct method is to normalise across all parent capsules that receive input from a particular child capsule, which will include normalising across parent capsule types and parent capsule positions. 

		We found this important detail somewhat tricky to implement, so we describe our implementation below. 

\begin{figure}[htbp]
			\centerline{\includegraphics[width=1\columnwidth]{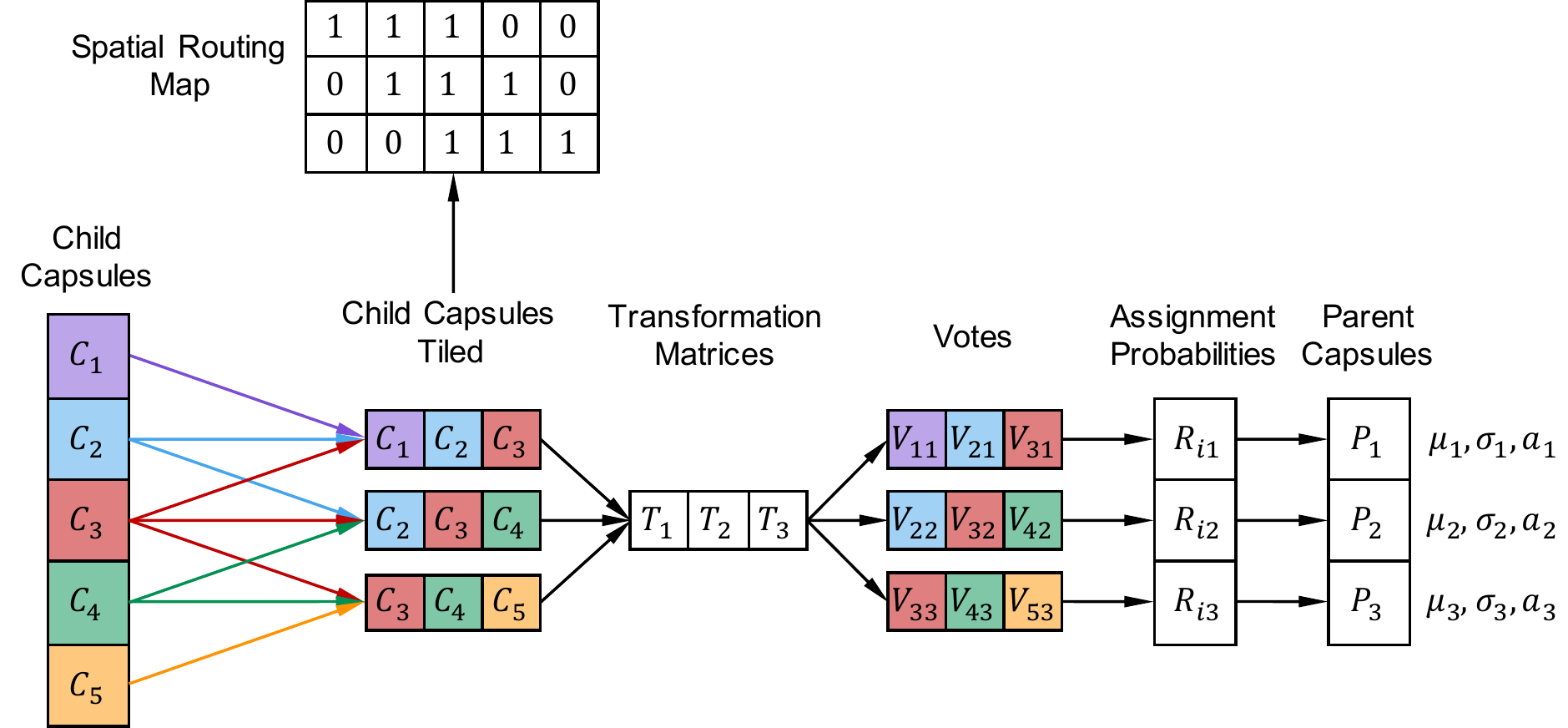}}
			\caption{Votes are computed by tiling the child capsules feeding to each parent capsule, and multiplying by the transformation matrix. The \textit{spatial routing map} is a binary matrix which stores the spatial connectivity between child capsules and parent capsules resulting from the convolution operation. $V_{ij}$ denotes the vote from child capsule $i$ to parent capsule $j$. The votes are scaled by the corresponding assignment probabilities $R_{ij}$ and used in the M-step to calculate the mean $\mu_{j}$, standard deviation $\sigma_j$, and activation $a_{j}$ of each parent capsule.}
			\label{fig:votes}
		\end{figure}

		With reference to Fig.~\ref{fig:votes}, the first step in computing the votes is tiling the child capsules according to the convolution kernel. The subsequent mapping between child capsules and parent capsules is stored a 2D binary matrix called the \textit{spatial routing map}. The tiled representation is then multiplied by a tensor containing $K$ transformation matrices ($3$ in our example), which are learned discriminatively with backpropagation during training. The tiled child capsules at each spatial location are multiplied by the same transformation matrices. The votes are then scaled by the corresponding assignment probabilities $R_{ij}$ and used in the M-step to compute the mean $\mu_{j}$, standard deviation $\sigma_j$, and activation $a_{j}$ of each parent capsule.

\begin{figure}[htbp]
			\centerline{\includegraphics[width=1\columnwidth]{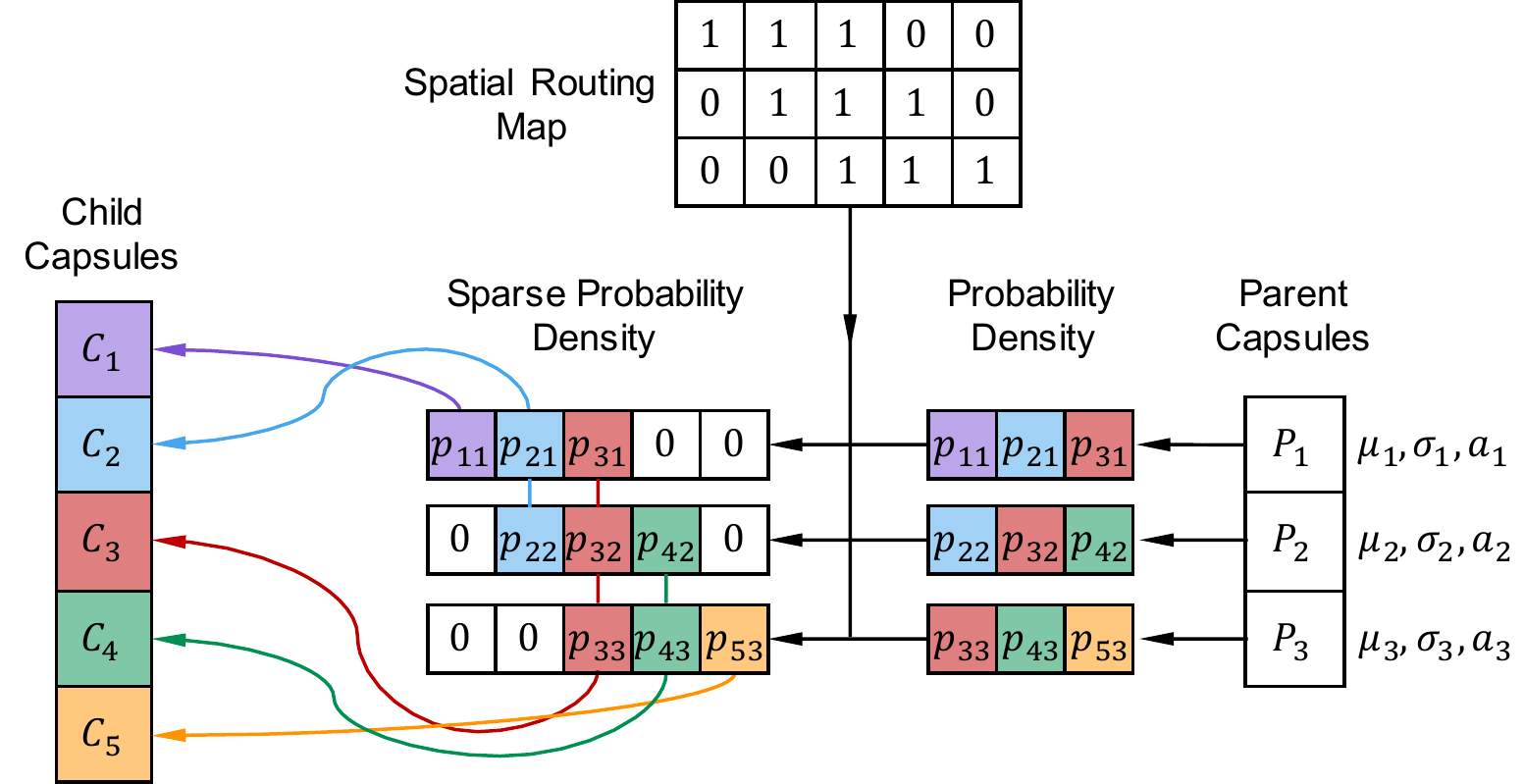}}
			\caption{Feedback of parent capsules to child capsules in the E-step of EM routing (follow diagram from right to left). $p_{ij}$ denotes the probability density of vote $V_{ij}$ under the Gaussian distribution of the parent capsule $j$. The \textit{spatial routing map}, which stores the spatial connectivity between child capsules and parent capsules, is used to convert the probability densities to sparse representation thereby aligning by child capsule. Finally, the probability densities are used to update the routing assignments $R_{ij}$ by normalising over columns of the sparse representation.}
			\label{fig:sparse}
		\end{figure}

		Our implementation of the E-step is shown in Fig.~\ref{fig:sparse}. The probability density $p_{ij}$ of vote $V_{ij}$ is computed from the Gaussian distribution of parent capsule $P_{j}$. The probability densities are then converted to sparse representation using the \textit{spatial routing map} that was stored during the convolution operation. In the sparse representation the probability densities of each child capsule are aligned in one column, thereby enabling us to normalise over all parent capsules competing for a child capsule $C_{i}$.   

		% \vspace{6pt}
		\subsubsection*{Extending from 1D to 2D Convolution} 

\begin{figure}[htbp]
			\centerline{\includegraphics[width=1\columnwidth]{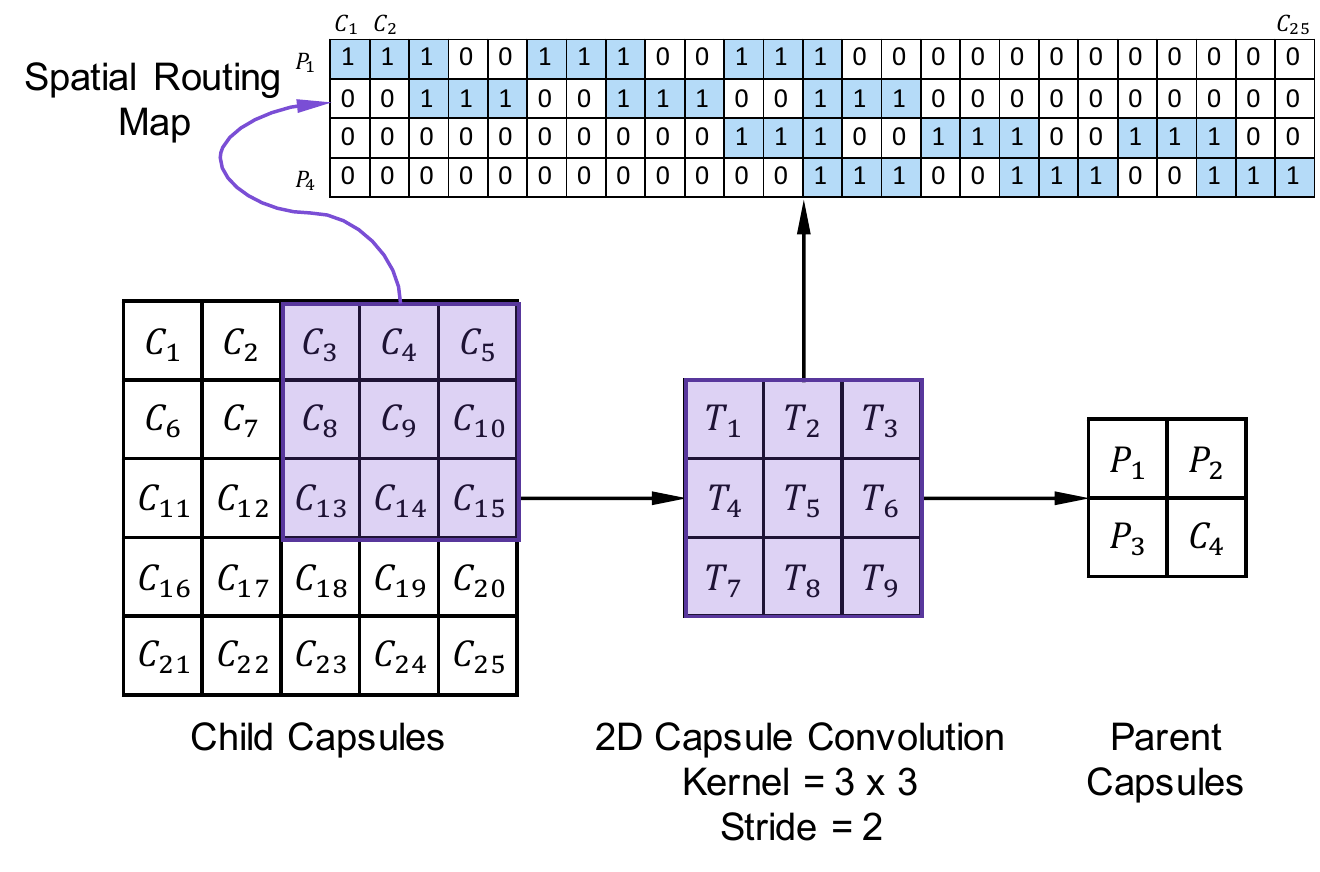}}
			\caption{Example of \textit{spatial routing map} produced by 2D capsule convolution with a 3x3 kernel and stride of 2.}
			\label{fig:2d_map}
		\end{figure}

		The above description of our implementation refers to the case of 1D capsule convolution, with a stride of 1. In order to extend to 2D convolution, we unroll the spatial dimension of the child capsules and parent capsules. Fig.~\ref{fig:2d_map} shows an example of the \textit{spatial routing map} produced from 2D capsule convolution with a 3x3 kernel and of stride 2. The rows of the \textit{spatial routing map} correspond to parent capsules, and the columns correspond to child capsules.

\section{Experiments}

	We implement ``Matrix Capsules with EM Routing'' by Hinton \textit{et al.}~\cite{Hinton2018} in TensorFlow, and test the \textit{smaller} capsule network configuration ($A=64$, $B=8$, $C=D=16$) on the smallNORB~\cite{LeCun2004} benchmark. We follow hyperparameter suggestions of the authors~\cite{Hinton2018a} and use a weight decay of $2 \times 10^{-7}$, and a learning rate of $3 \times 10^{-3}$ with exponential decay: $\textrm{decay steps}=2000$, $\textrm{decay rate}=0.96$. Lambda is set as follows \cite{Hinton2018b}: 
	$$ \lambda = 0.01*(1- 0.95^{i + 1})$$
	where $i$ is the routing iteration number (e.g. 0--2). The schedule for the increasing the margin in the spread loss is set as follows~\cite{Hinton2018b}:
	$$ margin = 0.2 + 0.79*sigmoid\big(min(10, step/50000 -4)\big) $$
	where $step$ is the training step. The batch size is set to $64$.

	% TABLE: Comparison to other open source implementations
	\begin{table}[!hb]
		\centering
		\caption{Comparison of test accuracy on smallNORB dataset for different implementations of ``Matrix Capsules with EM Routing'' by Hinton \textit{et al.}~\cite{Hinton2018}. For the open source implementations on GitHub, the test accuracy is reported as at 28/05/2019, and the specific commit is noted in the reference.}
		\label{tab:comparison}

		\begin{tabular}[t]{l l c c}
			\toprule
	 		Implementation	& Framework & Routing iterations & Test accuracy \\
			\midrule
				% myarrow~
				Hinton \cite{Hinton2018}		& Not available & 3 & \textbf{97.8}\% \\
				yl-1993 \cite{Lei2019}			& PyTorch	 	& 1 & 74.8\%  \\
				yl-1993 \cite{Lei2019}			& PyTorch	 	& 2 & 89.5\%  \\
				yl-1993 \cite{Lei2019}			& PyTorch	 	& 3 & 82.5\%  \\
				www0wwwjs1 \cite{Zhang2019} 	& Tensorflow 	& 2 & 91.8\%  \\
				Officium \cite{Huang2019}		& PyTorch 		& 3 & 90.9\%  \\
				Ours 							& TensorFlow 	& 1 & 86.2\%  \\
				Ours 							& TensorFlow 	& 2 & \textbf{95.4\%}  \\
				Ours							& TensorFlow 	& 3 & 93.7\%  \\
			\bottomrule
		\end{tabular}
	\end{table}	
	\begin{figure}[!hb]
		\centerline{\includegraphics[width=0.90\columnwidth]{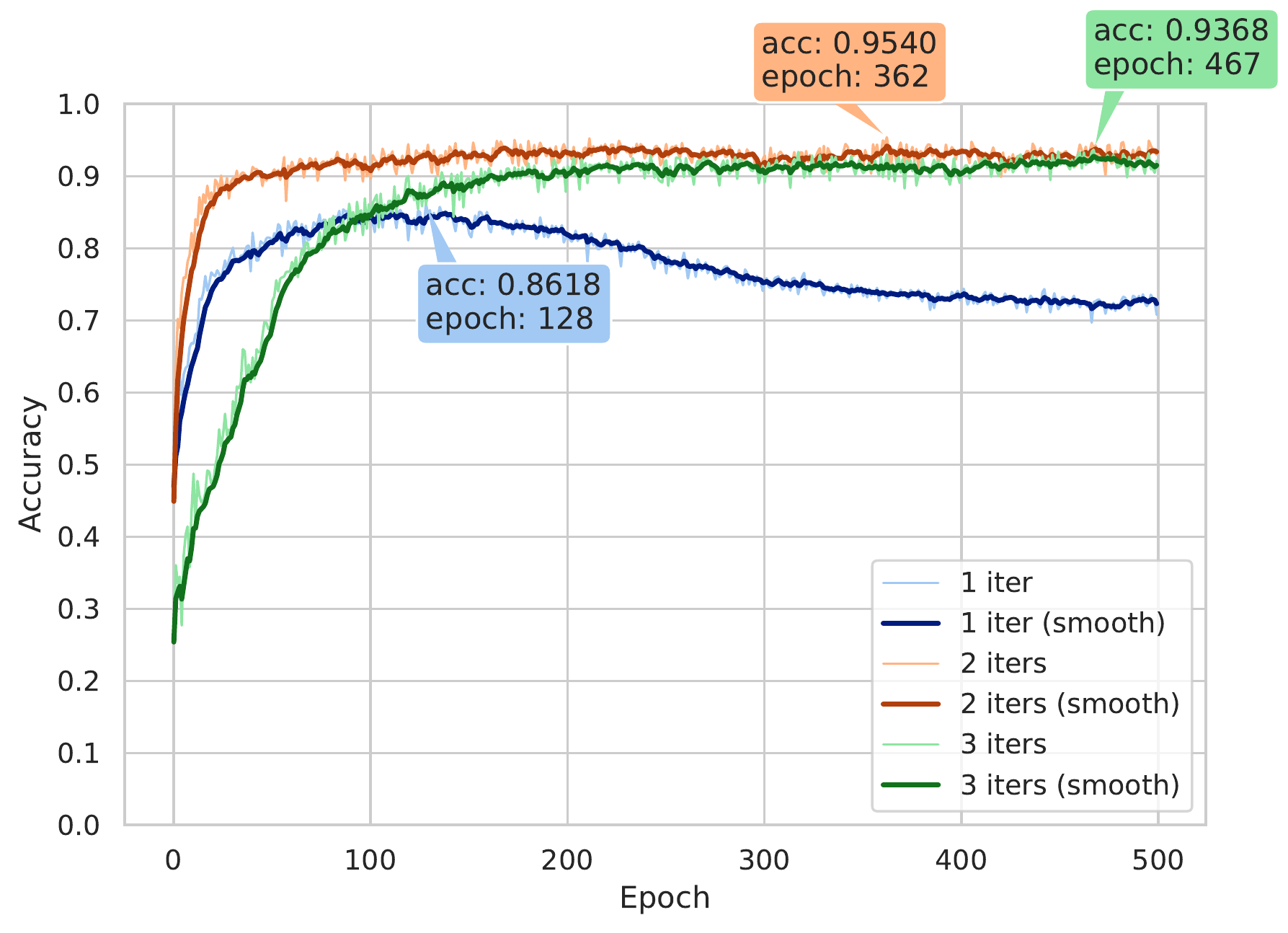}}
		\caption{Test accuracy of our implementation with 1--3 iterations of EM routing after each training epoch. Smoothed with exponentially-weighted moving window $\alpha = 0.25$.}
		\label{fig:val_acc}
	\end{figure}

	Fig.~\ref{fig:val_acc} shows the test accuracy of the matrix capsule model after each training epoch for 1--3 iterations of EM routing. Whereas Hinton \textit{et al.}~\cite{Hinton2018} report the maximum test accuracy at $3$ routing iterations, in our implementation the maximum test accuracy of $95.4\%$ occurs with $2$ routing iterations, and with $3$ iterations we record an accuracy of $93.7\%$.

	In Table~\ref{tab:comparison} we compare our implementation to other open source implementations available on GitHub. The accuracy of our implementation at $95.4\%$ is a $3.8$ percentage point improvement on the previous best open source implementation by Zhang (www0wwwjs1)~\cite{Zhang2019} at $91.8\%$, however it is still below the accuracy reported in Hinton \textit{et al.}~\cite{Hinton2018}. At this time, our implementation is currently the best open source implementation available.

\section{Conclusion}
	
	In this paper we discuss three common pitfalls when implementing ``Matrix Capsules with EM Routing'' by Hinton \textit{et al.}, and how to avoid them.  While our implementation performs considerably better than other open source implementations, nevertheless it still falls slightly short of the performance reported by Hinton \textit{et al.} (2018). The source code for this implementation is available on GitHub at the following URL: \url{https://github.com/IBM/matrix-capsules-with-em-routing}.  

% \clearpage

%
% ---- Bibliography ----
%
% BibTeX users should specify bibliography style 'splncs04'.
% References will then be sorted and formatted in the correct style.
%
\bibliographystyle{splncs04}
\bibliography{../../../../bib/ai}

\end{document}